\documentclass[conference]{IEEEtran}
\IEEEoverridecommandlockouts
\usepackage{amsmath,amssymb,amsfonts}
\usepackage{algorithmic,graphicx,textcomp,booktabs,xcolor,bm}
\usepackage[font=small]{subfig}
\usepackage{pgfplots,tikzscale}
\usepackage{balance,float}

\newcommand{\openness}{\mathit{openness}}
\newcommand{\score}{\mathit{score}}
\newcommand\TP{\mathit{TP}}
\newcommand\FN{\mathit{FN}}
\newcommand\TN{\mathit{TN}}
\newcommand\FP{\mathit{FP}}
\newcommand\BACC{\mathit{BACC}}

\renewcommand{\baselinestretch}{0.95}

\begin{document}
\title{Open-Set Recognition Using Intra-Class Splitting}

\author{
	\IEEEauthorblockN{Patrick Schlachter, Yiwen Liao and Bin Yang}
	\IEEEauthorblockA{Institute of Signal Processing and System Theory\\University of Stuttgart, Germany}
}

\maketitle

\begin{abstract}
This paper proposes a method to use deep neural networks as end-to-end open-set classifiers. It is based on intra-class data splitting. In open-set recognition, only samples from a limited number of known classes are available for training. During inference, an open-set classifier must reject samples from unknown classes while correctly classifying samples from known classes. The proposed method splits given data into typical and atypical normal subsets by using a closed-set classifier. This enables to model the abnormal classes by atypical normal samples. Accordingly, the open-set recognition problem is reformulated into a traditional classification problem. In addition, a closed-set regularization is proposed to guarantee a high closed-set classification performance. Intensive experiments on five well-known image datasets showed the effectiveness of the proposed method which outperformed the baselines and achieved a distinct improvement over the state-of-the-art methods.
\end{abstract}


\section{Introduction}
\begin{figure}[htbp]
	\centering
	\subfloat[Training]{\label{subfig:os-problem-training}\includegraphics[height=3.5cm,trim=0 0 8cm 1.1cm,clip]{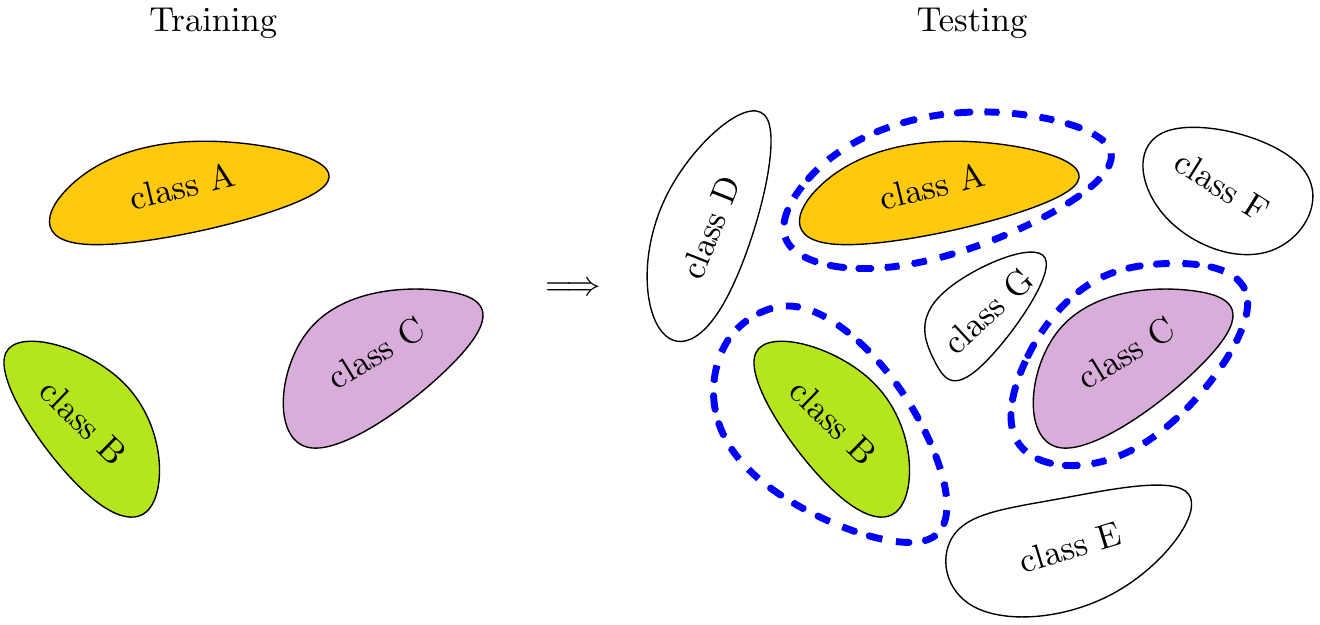}}
	\hfill
	\subfloat[Test]{\includegraphics[height=3.5cm,trim=6.5cm 0 0 1.1cm,clip]{figures/os-problem}}
	
	\caption{Open-set recognition: (a) Some known classes (class A, B and C) during training. (b) The unknown classes (class D, E, F and G) should be rejected while the known classes must be discriminated from each other by the learned decision boundaries (\emph{blue dashed frames}).}
	\label{fig:os-problem}
\end{figure}

In recent years, machine learning and deep learning have achieved a huge success in classification~\cite{lecun2015deep}. However, most approaches share the assumption that each sample during inference belongs to one of a fixed number of known classes. In other words, these models are trained and evaluated under a closed-set (CS) condition. Unfortunately, such a closed-set environment is ideal and not common in practice. Indeed, many real applications are subject to an open-set (OS) condition as shown in Fig.~\ref{fig:os-problem}, meaning that some test samples belong to classes that are unknown during training, so-called ``unknown unknowns''~\cite{Bendale2016,Geng2018}. For example, in the field of medical image classification, some test images may indicate a certain kind of disease which is unknown in advance. Such images should not be classified as any of the known classes but as belonging to a new abnormal class.

Although open-set recognition is a common scenario in practice, it lacks of attention in the past, because it is much harder to solve than closed-set problems. Conventional methods to OS problems are variants of the support vector machine (SVM) such as the 1-vs-set SVM~\cite{Scheirer_2013_TPAMI} or W-SVM~\cite{Scheirer_2014_TPAMIb}. However, as shown in~\cite{Scheirer_2014_TPAMIb}, they are sensitive to the thresholds for rejecting abnormal samples and therefore need abnormal samples to find a proper threshold during training, which is often not possible in practice. Moreover, these methods can only achieve a good performance with extracted features based on expert knowledge, which require a search and have a limited transferable performance. Thus, classical SVM-based methods only achieve a limited performance on complex datasets such as natural images~\cite{Scheirer_2014_TPAMIb}.

In contrast to conventional shallow models, deep neural networks such as VGG-16~\cite{Simonyan2014}, Inception~\cite{szegedy2015going} or ResNet~\cite{He2016} achieved state-of-the-art performance in classification and recognition. Moreover, a generative adversarial network (GAN) is able to generate more realistic images than ever~\cite{Goodfellow2014}. Intuitively, a modern approach to deal with open-set problems is to generate fake images based on a deep GAN and use them to model the abnormal class. Consequently, an open-set problem is reformulated to a closed-set classification problem~\cite{Neal2018}. However, GAN based methods bear the following challenges. First, the assumption that generated fake images can represent the unseen abnormal samples is not solid, because it is still an open question whether a GAN structure can really approximate the true data distribution~\cite{nalisnick2018deep}. Second, GANs tend to generate images indistinguishable from the majority of the training dataset. This is not desirable for open-set problems because discriminating those images from the original images leads to a poor closed-set accuracy. Finally, unknown abnormal samples cannot be compactly defined without any prior information as in~\cite{Neal2018}.
Hence, defining a reasonable objective to train a GAN-based OS classifier remains challenging.   

\begin{figure}
	\centerline{\includegraphics[width=\linewidth]{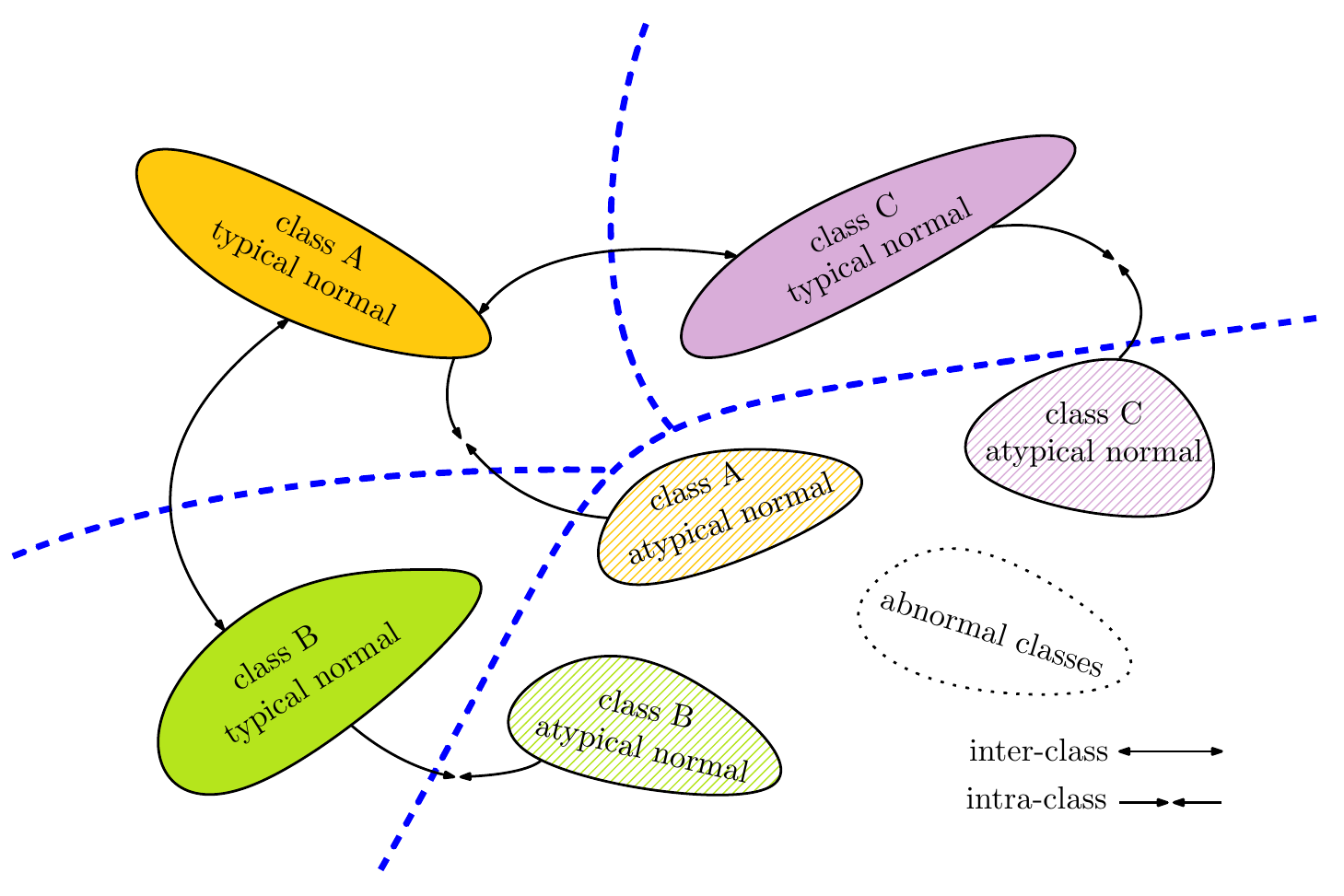}}
	\caption{The basic idea of the proposed method:
		The given normal dataset (class A, B and C) is split into typical and atypical normal subsets. The atypical normal subsets of all known classes are used to model the unknown abnormal classes and these subsets share the same new label during training. By discriminating the atypical normal samples from the typical normal ones, the trained classifier is expected to reject the abnormal classes as well.
	}
	\label{fig:splitting}
\end{figure}

Although modeling the unknown abnormal samples by generated fake images has some disadvantages, transforming an OS problem into a classification problem by introducing one additional class for all unknown classes still has a high potential for OS problems. A more natural idea to model abnormal samples is to use a certain part of the given normal samples. Schlachter et al. proposed an intra-class data splitting method which splits the given normal dataset into typical and atypical normal subsets and uses the latter to model the unknown abnormal class~\cite{Schlachter2019}. However, this method was originally designed for one-class classification. Therefore, it does not consider the inter-class information in OS problems, meaning the relations among the several known classes as shown in Fig.~\ref{fig:splitting}.

In this paper, we propose a novel deep learning method for open-set recognition problems based on improved intra-class splitting of data. In particular, a given $N$-class normal dataset is split into typical and atypical normal subsets. Then, the atypical normal samples are used to model the unknown abnormal data. Correspondingly, an OS problem is transformed into an $(N+1)$-class classification problem. In order to maintain a high closed-set classification ability, a novel closed-set regularized deep neural network is designed for this $(N+1)$-class classification.

Compared to prior work towards open-set problems, our work has three main contributions:
\begin{itemize}
	\item It is the first work using a small part of the given normal classes to model the unknown abnormal class. Accordingly, only the given normal samples are used during training without generating new fake samples. Therefore, no strong assumptions about the unknown abnormal samples are required. This is helpful for real-world open-set recognition scenarios.
	\item An improved intra-class splitting method is adapted to open-set recognition problems, which exploits the inter-class information among the given normal classes. The ameliorated splitting method uses the metric of class probability instead of the structural similarity index (SSIM)~\cite{Schlachter2019}. Hence, the new splitting method is more general and follows the human understanding as shown in Fig~\ref{fig:ty_aty_examples}.
	\item We propose a closed-set regularized deep neural network which realizes a high closed-set accuracy while having the ability of rejecting unknown abnormal samples.
\end{itemize}

\section{Proposed method}
\subsection{Basic idea}
The proposed method reformulates an original $N$-class open-set recognition problem into an $(N+1)$-class classification problem. This reformulation is realized by modeling the unknown abnormal class by atypical normal samples obtained through an improved intra-class data splitting.
Formally, a given set of samples $\mathcal{X}_i$, where $i$ indicates any of the known $N$ classes, is split into typical and atypical normal subsets $\mathcal{X}_{i, \mathrm{typical}}$ and $\mathcal{X}_{i, \mathrm{atypical}}$ as illustrated in Fig.~\ref{fig:splitting}. Then, the atypical normal subsets of all $N$ classes are considered as one additional class by assigning them a new label during training.

\begin{figure}[htbp]
	\centerline{\includegraphics[width=\linewidth]{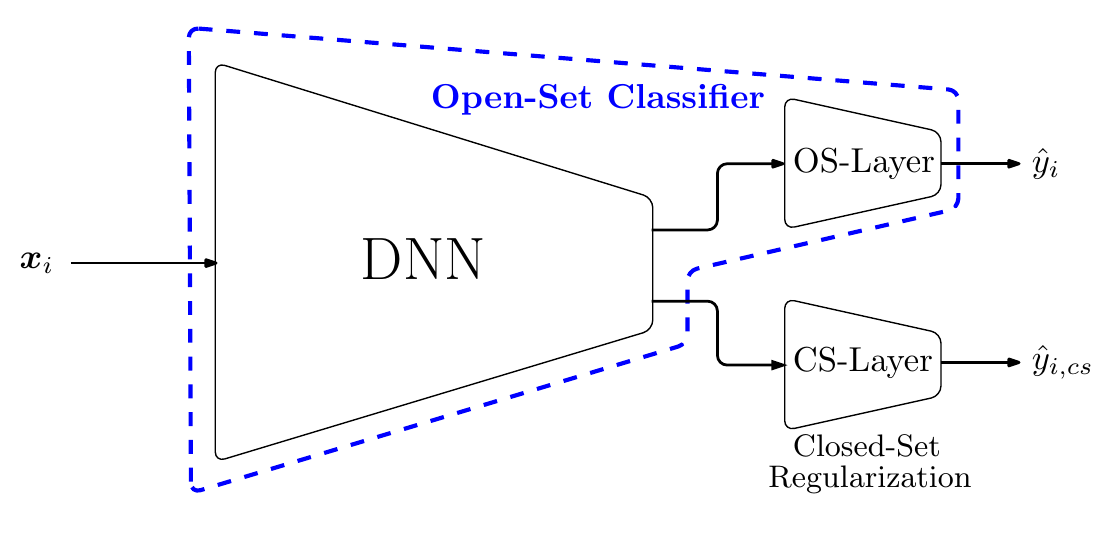}}
	\caption{The architecture of the proposed method. The open-set classifier is based on an arbitrary deep neural network (DNN) with $N+1$ outputs. The closed-set regularization is modeled by a one-layer network with $N$ outputs in this work.}
	\label{fig:cnn}
\end{figure}

Based on splitting given normal data, a deep neural network (DNN) with $N+1$ output neurons can then serve as an open-set classifier.	However, the atypical normal samples are actually normal samples and their new labels differ from the ground truth. Hence, a naive neural network with $N+1$ outputs will result in a low closed-set accuracy, because the atypical normal samples are incorrectly predicted. To prevent this situation, we propose a closed-set regularization subnetwork which forces the atypical normal samples to be correctly classified during training. Fig.~\ref{fig:cnn} visualizes the resulting architecture.

During inference, only the deep neural network without the closed-set regularization layer is used as an end-to-end classifier for open-set recognition which outputs a predicted label $\hat{y}_i$ for each input $\bm{x}_i$.

\subsection{Improved Intra-Class Data Splitting}
The original intra-class splitting method~\cite{Schlachter2019} trains an autoencoder and uses the reconstruction error of samples as a similarity score to split a given normal dataset. In particular, the samples with lower reconstruction errors are considered as typical normal, whereas the samples with higher reconstruction errors are atypical normal. This method works well for one-class classification problems. However, directly applying this autoencoder-based intra-class splitting method to OS problems is not optimal, because the available inter-class information is not utilized. More precisely, the discriminations among known classes are overlooked during splitting. 

In order to take full advantage of the inter-class information in OS problems, we use a multi-class classifier instead of an autoencoder for intra-class data splitting. Concretely, an $N$-class classifier is trained with the given $N$-class normal data. Once the classifier is trained, the incorrectly predicted samples and the correctly predicted samples with a low probability are selected as atypical normal samples. Thereby, probabilities correspond to the linear activations (logits) of the last layer in a regular deep neural network.

In general, this improved intra-class data splitting method is formulated as follows. Let $f(\cdot)$ indicate the mapping of an $N$-class neural network. $\bm{x}$ denotes a sample from the training dataset. Therefore, the predicted class probabilities under the learned mapping are $\hat{\bm{y}}_\text{prob} = f(\bm{x})$ with $\hat{\bm{y}}_\text{prob}\in \mathbb{R}^{N\times 1}$. Correspondingly, $\hat{\bm{y}}$ is the resulting class prediction in one-hot coding. Furthermore, let $\bm{y}\in \mathbb{R}^{N\times 1}$ be the ground truth in one-hot coding and $\odot$ be the element-wise product. Consequently, the score for intra-class splitting is denoted as
\begin{equation}
\score = (\hat{\bm{y}}_\text{prob}\odot \hat{\bm{y}}\odot \bm{y})^\top\cdot\bm{1}~,
\end{equation}
where $\score\in\mathbb{R}$ and $\bm{1}\in\mathbb{R}^{N\times 1}$ is a vector of ones. According to a predefined ratio $\rho$, the $\rho\%$ samples with the lowest scores are considered as atypical normal samples. The remaining samples are considered to be typical normal.

\begin{figure}[t]
	\centering
	\subfloat[typical normal samples]{\includegraphics[width=.48\linewidth]{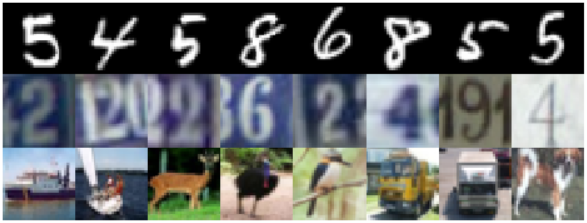}}
	\hfill
	\subfloat[atypical normal samples]{\includegraphics[width=.48\linewidth]{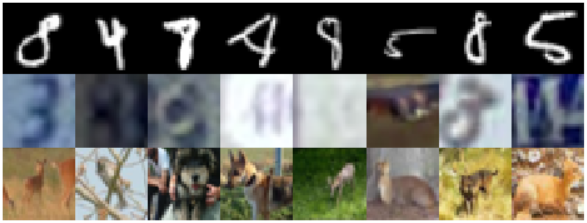}}
	\caption{Examples for typical and atypical samples according to intra-class data splitting.}
	\label{fig:ty_aty_examples}
\end{figure}

Therefore, the improved intra-class splitting method has two advantages:
\begin{itemize}
	\item The improved method is more general than the autoencoder-based one. Indeed, the original intra-class splitting was limited to image datasets due to utilizing SSIM as a similarity metric to split given normal data. In this work, the class probabilities are used as the metric to accomplish the intra-class data splitting. This is more general and can be extended to all kinds of datasets such as time series signals or extracted features.
	\item The inter-class information is taken into account. By training a multi-class classifier, only samples having low probability scores are selected as atypical normal samples. This splitting procedure matches the human understanding as shown in Fig.~\ref{fig:ty_aty_examples}.
\end{itemize}

\subsection{Closed-Set Regularization}
The neural network for $(N+1)$-class classification is an arbitrary regular deep neural network (DNN) with an additional subnetwork acting as a closed-set regularization as shown in Fig.~\ref{fig:cnn}. 

In this work, the additional closed-set regularization subnetwork only consists of one layer. Hence, the proposed architecture has two separate output layers: the OS-layer and the CS-layer. The OS-layer has $N+1$ neurons for open-set predictions. In contrast, the CS-layer has $N$ neurons and serves as a closed-set regularization layer. In particular, the OS-layer reserves an output neuron for the unknown abnormal class which is modeled by the atypical normal samples during training. On the other hand, in order to maintain a high closed-set accuracy as explained above, the CS-layer works as a regularization to force the atypical normal samples to be correctly classified to their own classes.

Consequently, the objective of the $(N+1)$-class neural network is learning to classify the training samples into the $N+1$ classes under the constraint that all training samples are still able to be classified into the given known $N$ classes using a simple one-layer subnetwork, see below.

\subsection{Loss Functions}
The objective of the entire network is transformed into a joint optimization problem and consists of two individual loss terms for the OS-layer and CS-layer as
\begin{equation}
\mathcal{L} = \mathcal{L}_\text{os} + \gamma\cdot\mathcal{L}_\text{cs}~,
\end{equation}
where $\mathcal{L}_\text{os}$ is the loss function for the OS-layer and $\mathcal{L}_\text{cs}$ is the loss function for the CS-layer. $\gamma$ is a hyperparameter to tune the ratio between these two terms.

Let $B$ be the minibatch size during training. Moreover, $1_{y_i\in y^{(n)}}$ is an indicator function which returns 1 if a given sample $\bm{x}_i$ with a scalar label $y_i$ belongs to the class $y^{(n)}$ and otherwise returns 0. Based on these notations, the two loss terms are introduced as:

\paragraph{OS-Loss} The open-set problem is transformed into an $(N+1)$-class classification problem due to the intra-class splitting. Therefore, the OS-loss is a simple $(N+1)$-class categorical cross-entropy loss
\begin{equation}
\mathcal{L}_\text{os} = -\frac{1}{B}\sum_{i=1}^{B}\sum_{n=1}^{N_\text{os}}1_{y_i\in y^{(n)}}\log[P(\hat{y}_i\in y^{(n)})]~,
\end{equation}
where $N_\text{os} = N+1$ and $P(\hat{y}_i\in y^{(n)})$ denotes the predicted probability that sample $\bm{x}_i$ belongs to the class $y^{(n)}$, meaning the value of the $n$-th element of the output vector of the network.
\paragraph{CS-Loss} The closed-set regularization loss is an $N$-class categorical cross entropy loss
\begin{equation}
\mathcal{L}_\text{cs} = -\frac{1}{B}\sum_{i=1}^{B}\sum_{n=1}^{N_\text{cs}}1_{y_i\in y^{(n)}}\log[P(\hat{y}_i\in y^{(n)})]~,
\end{equation}
where $\mathcal{L}_\text{cs}$ shares the same notation as $\mathcal{L}_\text{os}$ and $N_\text{cs}=N$ is the number of the given known classes.

\section{Experiments}
\subsection{Setup}
As a basic experiment, the proposed method was first evaluated on MNIST~\cite{lecun1998gradient}, SVHN~\cite{Netzer2011} and CIFAR-10~\cite{krizhevsky2009learning}. Each dataset has 10 classes and 6 of them were randomly selected as the known classes during training. Certainly, the test set consisted of the known 6 classes and the left 4 unknown classes. We repeated this basic experiment with 5 different seeds, i.e. 5 random combinations of the known classes. In order to evaluate the robustness of the proposed method over different openness, the second experiment was to train the model with 6 known classes in CIFAR-10 and test it with different numbers of unknown classes from CIFAR-100~\cite{krizhevsky2009learning} and Tiny ImageNet~\cite{ILSVRC15}. Finally, as $\rho$ is the key hyperparameter in our method, the sensitivity to $\rho$ was further evaluated with the same settings as the basic experiment except that 6 different $\rho$ were tested.

Balanced accuracy~\cite{brodersen2010balanced} was selected as the primary metric for evaluation, because it allows a fair comparison of balanced and imbalanced datasets which can both occur in OS problems. In this work, the unknown or abnormal classes are considered as negative while the known classes are considered as positive. Consequently, the balanced accuracy for OS problem is defined as
\begin{equation}
\BACC = \frac{1}{2}\cdot\left(\frac{\TP}{\TP+\FN} + \frac{\TN}{\FP+\TN}\right)~,
\end{equation}
where $\TP$ is the number of ``true positives". In contrast to binary problems, $\TP$ represents those samples which are correctly classified as one of the known classes and not only samples that are correctly classified as positive.\footnote{Correspondingly, $\FN$ are ``false positives", $\TN$ are ``true negatives" and $\FP$ are ``false positives".}

We selected the following four baseline models including one state-of-the-art method based on the generation of counterfactual images from the literature:
\begin{enumerate}
	\item WSVM: Weibull support vector machine with the default settings of the libsvm-openset package~\cite{Scheirer_2011_TPAMI}.
	\item OCSVM: An $N$-class network was trained for closed-set prediction, while a separate one-class SVM~\cite{scholkopf2001estimating} with $\nu=0.1$ was trained on the training dataset for rejecting abnormal samples. The final results were the multiplication of the predictions from both classifiers.
	\item GAN: $(N+1)$-class neural network using fake images as abnormal class which are generated by a regular GAN.
	\item CF: Same settings with counterfactual image generation method in~\cite{Neal2018}.
\end{enumerate}
Furthermore, two variants of the proposed method were evaluated to judge the effectiveness of closed-set regularization and improved intra-class data splitting:
\begin{enumerate}
	\setcounter{enumi}{4}
	\item NN-ics: An $(N+1)$-class neural network combined with the intra-class data splitting method but without any closed-set regularization layers.
	\item AE-ics: Same settings with the proposed method except that an autoencoder was used for intra-class data splitting as in~\cite{Schlachter2019}.
\end{enumerate}
Note that the baselines OCSVM, GAN, NN-ics and AE-ics shared the same architecture with the proposed method for a fair comparison and they were implemented by scikit-learn~\cite{scikit-learn} and TensorFlow~\cite{abadi2016tensorflow}.

The proposed method used a modified VGG-16~\cite{Simonyan2014} as a backbone with residual blocks~\cite{He2016} to reduce the number of network parameters. L2-regularization was used for each convolutional layer with a decay of $10^{-3}$. $\gamma$ was equal to 1 for the entire loss function. The splitting ratio $\rho$ was selected as 10 for MNIST and 20 for SVHN and CIFAR-10. Finally, the batch size was 32 and the model was trained for 50 epochs.

\subsection{Basic Experiments}
The basic experimental results are listed in Table~\ref{tab:baccu_vs_baseline}. The proposed method outperformed other baselines including the state-of-the-art methods in all conducted experiments. 

CIFAR-10, as a natural image dataset, is challenging in OS problems. The conventional shallow model WSVM or even the state-of-the-art method CF only reached a balanced accuracy of about 50\%. In comparison, our method achieved a balanced accuracy of more than 71\%, which corresponds to an improvement of 39\% over the other considered methods.

Considering the less difficult image datasets, MNIST and SVHN, both shallow and deep models showed a good performance. However, the proposed method still had the best performance with about 8\% higher balanced accuracy.

Interestingly, there was a huge gap between the performances of a regular GAN and that of CF. This showed the difficulty in designing a correct objective for generating fake samples to represent the unknown abnormal data, because there is no prior information of the unknowns during training. In contrast, our method only uses a part of the training dataset to model the abnormal samples which does not require any prior information.

Eventually, the proposed method with an improved intra-class splitting achieved a better performance than the baseline using autoencoder-based splitting as expected.
\begin{table}[ht]
	\centering
	\caption{Balanced Accuracy (standard deviation) in \%.}
	\label{tab:baccu_vs_baseline}
	\resizebox{\linewidth}{!}{
		\begin{tabular}{lcccccc}
			\toprule
			\multicolumn{1}{c}{\bf Dataset}  &\multicolumn{1}{c}{\bf WSVM}  &\multicolumn{1}{c}{\bf OCSVM}  &\multicolumn{1}{c}{\bf GAN}  &\multicolumn{1}{c}{\bf CF} &\multicolumn{1}{c}{\bf AE-ics} &\multicolumn{1}{c}{\bf Ours} \\
			\midrule
			MNIST & 84.6 {\scriptsize($\pm$3.5)} & 64.5 {\scriptsize($\pm$5.0)} & 55.6 {\scriptsize($\pm$2.6)} & 87.5 {\scriptsize($\pm$2.0)} & 82.7 {\scriptsize($\pm$2.5)} & \textbf{94.3} {\scriptsize($\pm$0.4)} \\	
			\midrule                                                                                                   
			SVHN & 75.2 {\scriptsize($\pm$3.3)} & 49.2 {\scriptsize($\pm$0.6)} & 48.4 {\scriptsize($\pm$1.1)} & 76.2 {\scriptsize($\pm$4.6)} & 72.2 {\scriptsize($\pm$3.3)} & \textbf{82.8} {\scriptsize($\pm$0.5)} \\
			\midrule	    	                                                                                       
			CIFAR-10 & 46.5 {\scriptsize($\pm$4.1)} & 50.0 {\scriptsize($\pm$3.0)} & 43.5 {\scriptsize($\pm$4.0)} & 51.2 {\scriptsize($\pm$0.7)} & 50.2 {\scriptsize($\pm$4.8)} & \textbf{71.2} {\scriptsize($\pm$2.1)} \\
			\bottomrule
	\end{tabular}}
\end{table}

\subsection{Performance with Different Openness}
Following~\cite{Neal2018}, the openness of the OS problem is defined as
\begin{equation}
\openness = 1-\sqrt{\frac{N}{M}}~,
\end{equation}
where $N$ denotes the number of the known normal classes during training and $M$ is the total number of the encountered classes during testing, i.e. $M = N+N'$ with $N'$ denoting the number of unknown abnormal classes during testing. To test the proposed method's robustness to different openness, we used the following settings. First, $N=6$ for all cases. Moreover, the number of abnormal classes $N'$ during testing was chosen as 20, 30, 50, 100 and 200. In the former four cases, we trained the model on CIFAR-10 and tested it on CIFAR-100. In the last case, we trained the model on CIFAR-10 and tested it on Tiny ImageNet. The corresponding results are listed in Table~\ref{tab:baccu_vs_openness}. Here we only compare our method with the state-of-the-art method CF from the basic experiment and the variant NN-ics of the proposed method.

Regarding challenging natural image datasets such as CIFAR-100 and Tiny ImageNet, the proposed method outperformed CF in all cases with high balanced accuracies. 

Note that the variant of the proposed method NN-ics, i.e. a naive neural network with intra-class data splitting method, already performed better than CF in all considered cases. Interestingly, all three methods showed a stable performance with different openness.

\subsection{Sensitivity to the splitting ratios}
\begin{figure}[t]
	\centering
	\includegraphics[width=\linewidth]{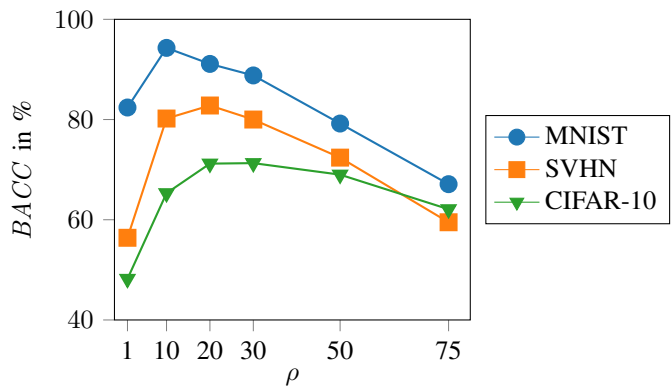}
	\vspace{-.5cm}
	\caption{The balanced accuracy vs. different splitting ratios $\rho$.}
	\label{fig:rho}
\end{figure}

The splitting ratio $\rho$ is a crucial hyperparameter for the proposed method. Fig.~\ref{fig:rho} shows the performance regarding different ratios. As expected, both a very low ratio and a very high ratio lead to a worse performance than proper ratios. A very low ratio, e.g. $\rho=1$, means that only a small part of the training data is used to represent abnormal data. Therefore, the training procedure is highly imbalanced and the trained model cannot gain adequate gradient information for the additional abnormal class during training. Consequently, the model can poorly identify abnormal samples during testing. On the other hand, a large ratio, e.g. $\rho=75$, causes too many normal samples to be incorrectly predicted as abnormal which results in a low closed-set accuracy.

From another perspective, the optimal $\rho$ is also an indicator for the homogeneity of a dataset. For instance, MNIST is more homogenous and hence requires less atypical samples than SVHN and CIFAR-10 which results in a smaller value for the optimal $\rho$.

Although $\rho$ plays an important role, our method is not very sensitive to this $\rho$ in a wide range. For example, as illustrated in Fig.~\ref{fig:rho}, the proposed method has a stable performance on SVHN with $\rho\in[10, 30]$. 

\section{Conclusion}
We proposed a novel deep learning method for open-set recognition. By using intra-class data splitting, it allows to introduce a categorical cross-entropy loss and additional closed-set regularizations. As a result, the proposed method allows end-to-end training of regular deep neural networks for open-set recognition. Our method was evaluated in a large amount of experiments with natural images. It showed a distinct improvement over state-of-the-art methods towards open-set recognition in average. Future work may integrate GANs into the proposed method to generate more atypical normal samples which further increase its performance and robustness. Furthermore, more realistic datasets such as fingerprints or face images will be used to evaluate the proposed method. Finally, we will also evaluate the proposed method on non-image datasets such as radar signals.

\balance
\begin{table}[H]
	\centering
	\caption{Balanced Accuracy vs. Openness in \%.}
	\label{tab:baccu_vs_openness}
	\resizebox{\columnwidth}{!}{
		\begin{tabular}{lccccc}
			\toprule
			\textbf{Method}  & $\bm{N'=20}$ & $\bm{N'=30}$ & $\bm{N'=50}$  & $\bm{N'=100}$ & $\bm{N'=200}$ \\
			\midrule
			CF		& 52.0 {\scriptsize($\pm$2.1)} & 53.2 {\scriptsize($\pm$2.0)} & 52.1 {\scriptsize($\pm$3.0)} & 52.6 {\scriptsize($\pm$3.1)} & 52.5 {\scriptsize($\pm$1.0)} \\
			\midrule
			NN-ics		& 57.1 {\scriptsize($\pm$2.3)} & 57.8 {\scriptsize($\pm$2.0)} & 57.7 {\scriptsize($\pm$1.9)} & 58.2 {\scriptsize($\pm$1.8)} & 58.4 {\scriptsize($\pm$2.5)} \\
			\midrule
			Ours	& \textbf{69.5} {\scriptsize($\pm$1.9)} & \textbf{70.0} {\scriptsize($\pm$1.9)} & \textbf{70.3} {\scriptsize($\pm$1.6)} & \textbf{70.8} {\scriptsize($\pm$1.5)} & \textbf{70.1} {\scriptsize($\pm$1.7)} \\
			\bottomrule
	\end{tabular}}
\end{table}

\bibliographystyle{IEEEtran}
\bibliography{refs}
\end{document}